\documentclass[runningheads]{llncs}
\usepackage[T1]{fontenc}
\usepackage{graphicx}
\usepackage{booktabs}
\usepackage[misc]{ifsym}

\usepackage{amsfonts}
\usepackage{amsmath}
\usepackage{amssymb}
\usepackage{theorem}
\usepackage{algorithm}
\usepackage{algpseudocode}
\usepackage{latexsym}
\usepackage{xcolor}
\newcommand{\dom}{{\rm dom}\,}
\newcommand{\argmin}{{\rm argmin}\,}
\newcommand{\etal}{et al.}

\newcommand{\1}{\mathbb{I}}

\begin{document}

\title{Online $\mathrm{L}^{\natural}$-Convex Minimization}

\author{Ken Yokoyama\inst{1} \and
Shinji Ito\inst{2} \and
Tatsuya Matsuoka\inst{3}\and  Kei Kimura\inst{1} \and Makoto Yokoo\inst{1}}
\institute{Kyushu University \email{yokoyama@agent.inf.kyushu-u.ac.jp,kkimura@inf.kyushu-u.ac.jp,yokoo@inf.kyushu-u.ac.jp}\and
NEC Corporation / RIKEN (His current affiliation is The University of Tokyo / RIKEN.) \email{shinji@mist.i.u-tokyo.ac.jp}\and
NEC Corporation \email{ta.matsuoka@nec.com}}
\authorrunning{Ken Yokoyama\inst{1} \and
Shinji Ito\inst{2} \and
Tatsuya Matsuoka\inst{3}\and  Kei Kimura\inst{1} \and Makoto Yokoo\inst{1}}
\maketitle            

\begin{abstract}
  An online decision-making problem is a learning problem in
  which a player repeatedly makes decisions in order to minimize the long-term
  loss. These problems that emerge in applications often have nonlinear combinatorial
  objective functions, and developing algorithms for such problems
  has attracted considerable attention. An existing general framework for
  dealing with such objective functions is the online submodular minimization.
  However, practical problems are often out of the scope of this
  framework, since the domain of a submodular function is limited to a subset
  of the unit hypercube. To manage this limitation of the existing framework,
  we in this paper introduce the online $\mathrm{L}^{\natural}$-convex minimization, where an
  $\mathrm{L}^{\natural}$-convex function generalizes a submodular function so that
  the domain is a subset of the integer lattice. We propose computationally
  efficient algorithms for the online $\mathrm{L}^{\natural}$-convex function minimization in
  two major settings: the full information and the bandit settings. We
  analyze the regrets of these algorithms and show in particular that our
  algorithm for the full information setting obtains a tight regret bound
  up to a constant factor. We also demonstrate several motivating examples that
  illustrate the usefulness of the online $\mathrm{L}^{\natural}$-convex minimization.

\end{abstract}

\section{Introduction}
Online decision-making is a learning problem in which a player
repeatedly chooses decisions and makes predictions for loss to minimize long-term loss.
These problems that appear as decision-making problems often have nonlinear and combinatorial (i.e., discrete) functions as the objective functions~\cite{pmlr-v206-tsuchiya23a,shalev2012online,qin2014contextual}.
Designing computationally efficient algorithms with low regret for such problems is challenging.

Problems with nonlinear combinatorial objective functions appear, for example, in price prediction optimization.
Price prediction optimization maximizes profits by predicting the demand for an unknown demand distribution and determining the prices of multiple items.
In practical applications, pricing is often discrete, such as offering discounts of 5\% or 10\%, and demand functions for pricing tend to be nonlinear.

An existing framework for dealing with such problems is online submodular minimization, where the objective function is submodular.
Submodularity, also known as the law of diminishing marginal utility, appears frequently  in various fields such as economics, machine learning, and operations research.

Although online submodular minimization has diverse applications, its  scope is also limited since the domain of the submodular function is a distributive lattice, which can be identified with a subset of the unit hypercube $\{0,1\}^d$.
In fact, in the above example, price prediction  optimization cannot be cast as an online submodular minimization if there are more than three types of items.

To overcome this limitation, we introduce online  $\mathrm{L}^{\natural}$-convex function minimization, where the minimizing objective function is an  $\mathrm{L}^{\natural}$-convex function. 
An $\mathrm{L}^{\natural}$-convex function is a generalization of a submodular function whose domain is a subset of the integer lattice~(i.e., $\mathbb{Z}^d$) rather than $\{0, 1\}^d$.
Therefore, online $\mathrm{L}^{\natural}$-convex function minimization can capture problems that are out of the scope of online submodular minimization.
Moreover, $\mathrm{L}^{\natural}$-convex functions are known to be transformed to multimodular functions via an unimodular transformation, and such multimodular functions appear in queueing theory, Malkov decision processes, and discrete event systems~\cite{altman2000multimodularity,altman2003discrete,freund2017minimizing}.
Hence, online $\mathrm{L}^{\natural}$-convex function minimization also captures online multimodular function minimization.

\subsection{Contributions}
In this paper, we propose algorithms for online $\mathrm{L}^{\natural}$-convex function minimization in two settings commonly addressed in previous studies.
The first one is the full information setting, where, after making a decision, the player has access to all information relevant to that decision.
The second one is the bandit setting, in which the player receives feedback only on the results of selected actions and cannot know the results of unselected actions.

We evaluate our algorithms in terms of regret, which is common in online decision making.
Regret $R_T$  is the difference between the sum of losses up to period $T$ in each iteration and the sum of losses for the fixed choice that is optimal in hindsight.
See Equation~\ref{eq_regret} in Section~\ref{section_preliminaries} for a formal definition of the regret.

Notation needed for explaining the contributions in this paper is listed in Table~\ref{TableNotation}.
The contributions of our study can be summarized as follows.
  \begin{table}[htb]
    \caption{Notation}
    \label{table_Notation}
    \centering 
    \begin{tabular}{ll}
      \toprule 
      Parameter & Meaning \\
      \midrule 
      $d\in \mathbb{Z}_+$ & Dimension of the decision space \\
      $[d]$ & $\{1, 2, \dots, d\}$ \\
      $T\in \mathbb{Z}_+$ & Time horizon \\
      $\check\gamma, \hat\gamma:[d]\rightarrow \mathbb{Z}$&Lower and upper bounds of the decision space\\
      $\mathcal{K}\subseteq \{z\in \mathbb{Z}^d \mid \check\gamma(i) \leq z_i \leq \hat\gamma(i)~(\forall i\in[d])\}$&Bounded decision space that is $\mathrm{L}^\natural$-convex set \\
      $f_t: \mathcal{K} \rightarrow [-M, M]$ & $\mathrm{L}^{\natural}$-convex cost function at time $t\in [T]$ \\
      $N$ & $\max_{i\in[d]}\{\hat \gamma(i)-\check\gamma(i)\}$ \\
      $\hat{L} \in \mathbb{R}$ & $l_\infty$-Lipschitz constant of $f_t~(\forall t\in[T])$ \\
      \bottomrule 
    \end{tabular}
    \label{TableNotation}
  \end{table}
\begin{itemize}
  \item    In the full information setting of online L$^{\natural}$-convex function minimization, we propose a computationally efficient
  randomized algorithm that  achieves the following regret bound:
  $$\mathbf{E}[R_T]=O(\hat{L}N\sqrt{dT}).$$
  \item In the bandit setting of online L$^{\natural}$-convex function minimization, we propose a computationally efficient randomized algorithm that  achieves the following regret bound:
  $$\mathbf{E}[R_T]=O(MdNT^{2/3}).$$
  \item In the Online L$^{\natural}$-convex function minimization, for any algorithm,
  there is a sequence of L$^{\natural}$-convex cost functions such that the algorithm has regret at least $\Omega (\hat{L}N\sqrt{dT})$.
  Therefore, our proposed algorithm for the full information setting achieves the best regret bound up to a constant factor.
  \item We also present an example of problems that can be naturally formulated as online $\mathrm{L}^{\natural}$-convex minimization in Section~\ref{application}.
\end{itemize}
Table~\ref{TableRegretBounds} summerizes regret bounds on the models of online submodular minimization and online $\mathrm{L}^{\natural}$-convex minimization.
  \begin{table}[htb]
    \caption{Regret bounds in online submodular and $\mathrm{L}^{\natural}$-convex minimization}
    \label{table_contribution}
    \centering 
    \begin{tabular}{lccc}
      \toprule
      Model & \begin{tabular}{l}Upper bound \\ (full-information)\end{tabular} & \begin{tabular}{l}Upper bound \\ (bandit)\end{tabular} & Lower bound \\
      \midrule
      Online submodular minimization & $O(\hat{L}\sqrt{dT})$ & $O(MdT^{2/3})$ & $\Omega (\hat{L}\sqrt{dT})$ \\
      Online $\mathrm{L}^{\natural}$-convex minimization & $O(\hat{L}N\sqrt{dT})$ & $O(MdNT^{2/3})$ & $\Omega (\hat{L}N\sqrt{dT})$ \\
      \bottomrule
    \end{tabular}
    \label{TableRegretBounds}
  \end{table}

\subsection{Related Work}
$\mathrm{L}^{\natural}$-convex functions are central functions in discrete convex analysis, which aims to establish a general framework for minimization of discrete functions (i.e., functions defined on the integer lattice) by means of a combination of the ideas in continuous and discrete optimization.
As mentioned above, $\mathrm{L}^{\natural}$-convex functions generalize submodular functions and can formulate various problems in diverse fields such as operations research~\cite{chen2017convexity,chen2021discrete}, economics~\cite{murota2022discrete}, and computer vision~\cite{shioura2017algorithms}.
A combination of $\mathrm{L}^{\natural}$-convex functions and machine learning have been seen in, e.g,  \cite{zhang2021stochastic,sakaue2023rethinking}.
An efficient algorithm has been proposed to minimize $\mathrm{L}^{\natural}$-convex functions~\cite{murota1998discrete}, however devising an algorithm for online $\mathrm{L}^{\natural}$-convex minimization requires a careful combination of online optimization and discrete convex analysis and thus it is a nontrivial task.

Compared with the number of online decision-making problems on continuous domains, the number of those on discrete domains are relatively small.
A submodular function is a discrete function which appears in a variety of applications in the field matching online optimization such as price optimization and thus online submodular optimization is a well-studied topic.
Note that submodular functions are special cases of $\mathrm{L}^{\natural}$-convex functions with the domain restricted to $\{ 0,1\}^d$.
For online submodular minimization, Hazan and Kale~\cite{hazan2012online} obtained a tight regret bound for the full information setting, while Bubeck \etal~\cite{bubeck2021kernel} obtained that for the bandit setting for general online convex function minimizatin and thus for online submodular minimization (through a convex extension).
Chen \etal~\cite{chen2018online} gave an algorithm for online continuous submodular maximization and demonstrate its performance in experiments.

We should note that our techniques for online $\mathrm{L}^{\natural}$-convex minimization resemble those for stochastic $\mathrm{L}^{\natural}$-convex minimization \cite{zhang2021stochastic}, however problem settings are different in that they aim to obtain PAC grarantees in stochastic models.
\section{Preliminaries and Problem Statement}\label{section_preliminaries}
In this section, we present the fundamental properties of L$^{\natural}$-convex functions, alongside the concepts of prediction and online convex optimization, before detailing the specific problem setting.

We suppose that the decision space of a player is an $\mathrm{L}^{\natural}$-convex set.
\begin{definition}[$\mathrm{L}^{\natural}$-convex set~\cite{fujishige2000notes}]
  A set $\mathcal{K}\subseteq \mathbb{Z}^d$ is an $\mathrm{L}^{\natural}$-convex set if it satisfies
  \begin{equation}
  p, q \in \mathcal{K} \Longrightarrow \left\lceil\frac{p+q}{2}\right\rceil,\left\lfloor\frac{p+q}{2}\right\rfloor \in \mathcal{K}.
  \end{equation}
\end{definition}  

Hereafter, $\mathcal{K}$ means an $\mathrm{L}^{\natural}$-convex set.
We assume that $\mathcal{K}$ is bounded throughout the paper.
We also assume, without loss of generality, that $\mathcal{K}$ is full-dimensional.
We further assume that cost functions are L$^{\natural}$-convex functions defined as follows:
\begin{definition}[$\mathrm{L}^{\natural}$-convex function~\cite{fujishige2000notes}]\label{l-convex}
  A function $f:\mathcal{K} \rightarrow \mathbb{R}$ is called an L$^{\natural}$-convex function if it satisfies the discrete midpoint convexity:
  \begin{equation}
    f(x)+f(y) \geq f\left(\left\lceil \frac{x+y}{2}\right\rceil\right)+f\left(\left\lfloor\frac{x+y}{2}\right\rfloor\right),\quad \forall x,y\in \mathcal{K},
  \end{equation}
  where $\lceil \cdot \rceil$, $\lfloor \cdot \rfloor$ are the ceiling and flooring functions applied componentwisely to vectors. 
  \end{definition}
 
  In our online decision-making problem, the goal of the player is to minimize the cumulative cost $\sum_{t=1}^T f_t(z_t)$.
 The preformance of the player is evaluated by means of the regret $R_T$ defined as
    \begin{equation}\label{eq_regret}
      R_{T}=\sum_{t=1}^{T}f_t(z_t)-   \min_{z^*\in \mathcal{K}}  \sum_{t=1}^{T}f_t(z^*).
    \end{equation}
  \subsection{Online L$^{\natural}$-Convex Minimization}
  In the online $\mathrm{L}^{\natural}$-convex minimization, across iterations $t = 1, 2, \dots, T$, an online decision maker is tasked with consistently determining the point $z_t \in \mathcal{K}$.
  Following the selection of $z_t$ in each iteration, feedback is provided in the form of the cost $f_t(z_t)$, where $f_t: \mathcal{K} \rightarrow \mathbb{R}$ is an $\mathrm{L}^{\natural}$-convex function.

  In this paper, we consider two distinct problem settings characterized by different levels of information feedback:
  \begin{itemize}
    \item In the full information setting, at each round $t$, the player has comprehensive access to  the sequence of past functions $f_1, f_2,\dots,f_{t-1}$ applicable to any input.
    \item In the bandit setting, at each round $t$, the player is restricted to observing the values of the functions $f_1(z_1), f_2(z_2), \dots, f_{t-1}(z_{t-1})$ corresponding to one's previous selections $z_j~(j=1, 2, \dots, t-1)$.
  \end{itemize}
  \subsection{Lov\'asz Extension of Submodular Functions}
  We introduce the Lov\'asz extension on submodular functions, a key algorithmic tool.
  We leverage the fact that the restriction of an $\mathrm{L}^{\natural}$-convex function on an arbitary unit hypercube $z+\{0, 1\}^d~(z \in \mathbb{Z}^d)$, results in a submodular function~\cite{murota1998discrete}.
  Further, for any $\mathrm{L}^{\natural}$-convex set $\mathcal{K}$ and $z \in \mathcal{K}$, $\mathcal{K} \cap (z + \{ 0, 1 \}^d )$ is a distributive lattice, where a submodular function can be defined.
  To define a convex extension of $\mathrm{L}^{\natural}$-convex set and function, we define the Lov\'asz extension of a submodular function on distributive lattice.
  Without loss of generality we assume that distribute lattices appear in this paper are simple.
\begin{definition}[simple distributive lattice]\label{def_simple_distributive_lattice}
  $D\subseteq \{0, 1\}^d$ is said to be a simple distributive lattice if it satisfied the following a condition:    
  \begin{equation}
        \text{there exists a poset}~\mathcal{P} = ([d], \preceq)~\text{such that}~D = \{I \subseteq [d]\mid \text{I is a lower ideal of }\mathcal{P}\}.
    \end{equation}
\end{definition}

This definition follows from Theorem 3.9 in~\cite{fujishige2005submodular}.
Let $\overline{D}$ denote the convex hull of $D$.
  Before we proceed with the formal definition of the Lov\'asz extension, we introduce the concept of a chain.
   \begin{definition}[chain]\label{chain}
    A chain is a collection of subsets $A_0, A_1,\dots, A_p$ of $[d]$ such that
    \begin{equation*}
      A_0 \subsetneq A_1 \subsetneq A_2 \subsetneq \dots \subsetneq A_p.
    \end{equation*}
  \end{definition}
  
  For any $x\in \overline{D}$, there exists a unique chain that can represent $x$ as a convex combination:
  \begin{equation}
  x = \sum_{i=0}^p \mu_i \chi_{A_i}
  \end{equation}
  where $\mu_i > 0,~\sum_{i=0}^{p}\mu_i = 1,~ \chi_{A_i} \in D$ is the characteristic vector of $A_i \subseteq [d]~(i=0, \dots, d)$~\cite{fujishige2005submodular}.
  
  It is known that any maximal chain such that all the characteristic vectors of the elements in the chain are in $D$ has length $d+1$ when $D$ is a simple distributive lattice.
 Moreover, such maximal chains have one-to-one correspondence to the set of total orders obtainable by topologically sorting the elements of poset $\mathcal{P}=([d],\preceq)$ representing $D$ as in Definition~\ref{def_simple_distributive_lattice}.

  We define the Lov\'asz extension on a simple distributive lattice.
    \begin{definition}[Lov\'asz extension on a simple distributive lattice~\cite{fujishige2005submodular}]\label{lovasz_extension}
      For a submodular function on $D$, its Lovasz extension $\hat{f}$ on $\overline{D}$ is defined as follows.
      Let $x\in \overline{D}$, and let $A_0\subsetneq A_1 \subsetneq A_2\subsetneq \dots \subsetneq A_p$
       be the unique chain  such that $x=\sum_{i=0}^{p}\mu_i \chi_{A_i},\,\mu_i >0,~\sum_{i=0}^{p}\mu_i = 1$.
      Then the value of the Lov\'asz extension $\hat f$ at $x$ is defined to be
      \begin{equation}
      \hat{f}(x):=\sum_{i=0}^{p}\mu_i f(\chi_{A_i}).
      \end{equation}
      \end{definition}
      
      We say a maximal chain is associated with $x$ if (i) all of its elements are contained in $D$ and (ii) it contains the unique maximal chain in Definition~\ref{lovasz_extension} as a subchain.
      We highlight the properties of the Lov\'asz extension and submodular functions that are particularly critical, as outlined below.
  \begin{lemma}[Properties of the Lov\'asz extension~\cite{bach2013learning,fujishige2005submodular}]\label{lem_lovasz}
    Let $f:D\rightarrow \mathbb{R}$ be a submodular function with $D$ being a simple distributive lattice and let $\hat{f}:\overline{D}\rightarrow \mathbb{R}$ be the Lov\'asz extention of $f$.
    The following properties hold for $\hat{f}$:
    \begin{itemize}
      \item $\hat{f}$ is a convex function.
      \item For any $x\in D$, it holds that $\hat{f}(x)=f(x)$.
      \item For $x \in \overline{D}$, choosing a threshold $\tau\in [0, 1]$ uniformly at random and defining the level set $S_{\tau}=\{i:x_i>\tau\}$, we obtain $\hat{f}(x)=\mathbf{E}_\tau[f(\chi_{S_\tau})]$.
      \item For $x\in \overline{D}$, and an arbitrary maximal chain $\emptyset = A_0 \subsetneq A_1 \subsetneq A_2\subsetneq \dots \subsetneq A_d=[d]$  associated with $x$,  a subgradient $g$ of $\hat f$ at $x$ is given by:
      \begin{equation}\label{eq_lovasz}
        g(A_i - A_{i-1}) = f(A_i) - f(A_{i-1}).
      \end{equation}
    \end{itemize}
  \end{lemma}

  \subsection{Convex Extension of $\mathrm{L}^{\natural}$-Convex Functions}
We introduce a convex extension of an $\mathrm{L}^\natural$-convex function by piecing together the Lov\'asz extension of submodular functions introduced in the previous subsection.
As a preparation, we introduce a maximal chain associated with $x\in \overline{\mathcal{K}}$.
\begin{definition}[maximal chain associated with $x$]\label{def_maximal_as_x}
  Let $x\in \overline{\mathcal{K}}$.
  We define $\emptyset = A_0 \subsetneq A_1 \subsetneq \dots \subsetneq A_d = [d]$ to be a maximal chain associated with $x$ if it satisfies:
  \begin{align*}
    \exists \underline{z}\in \mathcal{K}, \underline{z}\leq x\leq \underline{z}+\boldsymbol{1}, z+\chi_{A_{i}} \in \mathcal{K}~(\forall i\in\{0, 1, \dots, d\}),\\
    \exists \mu_i\geq 0~(\forall i\in \{0, 1, \dots, d\}), x=\underline{z}+\sum_{i=0}^d \mu_i \chi_{A_i}, \sum_{i=0}^d \mu_i = 1.
  \end{align*}
\end{definition}

We define a convex extension of $\mathrm{L}^{\natural}$-convex functions.
\begin{definition}[convex extension of $\mathrm{L}^{\natural}$-convex functions~\cite{fujishige2005submodular}]\label{def_convexextension_onL}
  Let $f:\mathcal{K}\rightarrow \mathbb{R}$ is an $\mathrm{L}^{\natural}$-convex function, and let $x \in \overline{\mathcal{K}}$.
  We define a convex extension $\hat f:\overline{\mathcal{K}}\rightarrow \mathbb{R}$ of $f$ as follows.
  Let $\emptyset =A_0 \subsetneq A_1 \subsetneq \dots \subsetneq A_d = [d], \underline{z}\in\mathcal{K}$ be a maximal chain associated with $x$, and $\underline{z}$ and $\mu_i$ be those appearing in Definition~\ref{def_maximal_as_x}.
  For  $x \in \overline{K}$, we define the convex extension $\hat f$ of $\mathrm{L}^{\natural}$-convex functions $f$ as follows:
  \begin{equation}\label{eq_convexextension_L-natural}
    \hat{f}(x):=  \sum_{i=0}^k \mu_i f(\underline{z}+\chi_{A_i}).
  \end{equation}
\end{definition}

This convex extension $\hat f$ is piecewise linear by definition, and the subgradient of $\hat f$ at $x\in \overline{\mathcal{K}}$ can be easily computed by the chain.

Next, we introduce an important property of $\hat f$ defined by Equation~\ref{eq_convexextension_L-natural}.
\begin{lemma}[\cite{fujishige2005submodular}]
  A function $f:\mathcal{K} \rightarrow \mathbb{R} $ is an $\mathrm{L}^{\natural}$-convex function if and only if its convex extension $\hat{f}$ defined in Definition~\ref{def_convexextension_onL} is a convex function.
\end{lemma}

In addition, as with the case of submodular function, the following is known.
\begin{lemma}
  Let $f:\mathcal{K} \rightarrow \mathbb{R}$ be an $\mathrm{L}^{\natural}$-convex function and $\hat f$ be the convex extension of $f$ defined in Definition~\ref{def_convexextension_onL}.
  Let $x \in \overline{\mathcal{K}}$, and let $\emptyset =A_0 \subsetneq A_1 \subsetneq \dots \subsetneq A_d = [d], \underline{z}\in\mathcal{K}$ be a maximal chain associated with $x$, and $\underline{z}$ and $\mu_i$ be those appearing in Definition~\ref{def_maximal_as_x}.
  Choosing a threshold $\tau\in [0, 1]$ uniformly at random and defining the level set $S_{\tau}=\{i:x_i>\underline{z}_i+\tau\}$, we obtain $\hat{f}(x)=\mathbf{E}_\tau[f(\underline{z} + \chi_{S_\tau})]$.
\end{lemma}
\begin{proof}
  First, the restriction of an $\mathrm{L}^{\natural}$-convex function on each hypercube $D \cap (z+\{0, 1\}^d)~(z\in\mathbb{Z}^d)$ is a submodular function~\cite{murota1998discrete}.
  Next, it is shown, following Lemma~\ref{lem_lovasz}, that the Lov\'asz extension on a submodular function coincides with the expected value under a uniform distribution.
  Therefore, it holds that $\hat{f}(x)=\mathbf{E}_\tau[f(\underline{z} + \chi_{S_\tau})]$.\qed
\end{proof}
\subsection{Subgradient of the Convex Extension of $\mathrm{L}^{\natural}$-Convex Function}
We have defined the convex extension $\hat{f}:\overline{\mathcal{K}} \rightarrow \mathbb{R}$ of an $\mathrm{L}^{\natural}$-convex function $f:\mathcal{K} \rightarrow \mathbb{R}$ in the previous subsection.
Our algorithms in Section 3 rely on the fact that the subgradient $g$ of $\hat{f}$ at a given point $x \in \overline{\mathcal{K}}$ can be computed efficiently.
Here, we explain how this can be done.

Recall that the convex extension $\hat{f}$ of an $\mathrm{L}^{\natural}$-convex function $f$ is constructed by piecing together the Lov\'asz extension of the submodular function restricted to each unit hypercube $z+\{0,1\}^d$ ($z \in \mathbb{Z}^d$).
Since we can compute the subgradient for submodular functions, it suffices to show that, given a point $x \in \overline{\mathcal{K}}$, we can compute a maximal chain associated with $x$~(see Definition~\ref{def_maximal_as_x}).
This can be easily done when each $x_i \notin \mathbb{Z}$, since we can set $z =\lfloor x \rfloor$ and express $x$ as a convex combination of the characteristic vectors corresponding to a maximal chain of $z+\{0,1\}^d$.
Hence, the challenge here is to compute such a maximal chain when $x_i \in \mathbb{Z}$ for some $i \in [d]$ (consider the case when $\mathcal{K}=\{ y \in \mathbb{Z}^2 \mid y_1,y_2 \le 2 \}$ and $x=(2,2)$, where we cannot set $z =\lfloor x \rfloor$), and we show how to do this in the following.

We use the fact that 
the domain $\mathcal{K}$ is an $\mathrm{L}^{\natural}$-convex set that can be expressed by a system of linear inequalities using $\check{\gamma},\hat{\gamma}:[d]\to \mathbb{Z}$ and $\gamma: [d]^2 \to \mathbb{Z}$ as follows (see, e.g., \cite{murota2003discrete}):
\begin{align}
\label{eq:L-convex-linear-ineq-expression}
\mathcal{K} = P(\gamma,\check{\gamma},\hat{\gamma}) \cap \mathbb{Z}^d, 
\end{align}
where 
\begin{align}
P(\gamma,\check{\gamma},\hat{\gamma}):= \{ x \in \mathbb{R}^d \mid \check{\gamma}_i \le x_i \le \hat{\gamma}_i~(\forall i\in [d]), x_i-x_j \le \gamma_{i,j} (\forall i,j \in [d]) \}.
\end{align}
It is also known that $\overline{\mathcal{K}} = P(\gamma,\check{\gamma},\hat{\gamma})$.
We assume that the expression~\eqref{eq:L-convex-linear-ineq-expression} is explicitly given.

As in the case of submodular functions, we assume without loss of generality that $\overline{\mathcal{K}}$ is full-dimensional~(this corresponds to that the domain of a submodular function is a simple distributive lattice).

As noted above, we would like to compute a maximal chain associated with $x$ for given $x \in \overline{\mathcal{K}}$.
As shown in Subsection 2.3, a maximal chain can be found when $(z+[0,1]^d) \cap \overline{\mathcal{K}}$ is full-dimensional.
Hence, above computation can be reduced to the following problem: 
given $x \in \overline{\mathcal{K}}$, find $\underline{z} \in \mathbb{Z}^d$ such that $(\underline{z}+[0,1]^d) \cap \overline{\mathcal{K}}$ is full-dimensional, i.e., a simple distributive lattice.
This problem can be solved by the following procedure.

\begin{algorithm}[htb]
  \caption{Compute a maximal chain in an $\mathrm{L}^{\natural}$-convex set}\label{alg_compute_chain}
  \begin{algorithmic}
  \Require $\mathrm{L}^{\natural}$-convex set $\mathcal{K}$ and $x\in \overline{\mathcal{K}}$
    \State $D_0 \leftarrow \overline{\mathcal{K}}$.
  \For{$i=1, 2, \dots, d$}
      \If{$x_i \notin \mathbb{Z}$}
          \State $\underline{z}_i \leftarrow \lfloor x_i \rfloor$.
      \Else
          \State $\delta_i \leftarrow \max_{y \in D_i} y_i$.
          \If{$\delta_i = x_i$}
              \State $\underline{z}_i \leftarrow x_i - 1$.
          \Else
              \State $\underline{z}_i \leftarrow x_i$.
          \EndIf
      \EndIf
      \State $D_{i} \leftarrow D_{i-1} \cap \{ y \in \mathbb{R}^d \mid \underline{z}_i \le y_i \le \underline{z}_i + 1 \}$.
  \EndFor
  \State $f_{\underline{z}} \leftarrow f|_{\mathcal{K}\cap (\underline{z}+[0,1]^d)}$ 
  \State /*Compute the poset representation $([d],R)$ of $\dom(f_{\underline{z}})$ as follows.*/
  \State $R \leftarrow \emptyset$ (as a binary relation).
  \For{each $i, j\in [d]$}
  \State $\delta_{i,j}:=\max_{y \in \overline{\mathcal{K}}} y_i-y_j$.
  \If{ $\underline{z}_i-\underline{z}_j = \delta_{i,j}$}
  \State $R\leftarrow R\cup\{(i, j)\}$.
  \EndIf
  \EndFor
  \State Topologically sort $([d],R)$ and compute a permutation $\pi: [d] \to [d]$ corresponding to this sorting.
  \State \Return $\underline{z}$ and a maximal chain $\emptyset = A_0 \subsetneq A_{1} \subsetneq \dots \subsetneq A_{d} = [d]$, where $A_i \leftarrow \{ \pi(1),\pi(2),\dots, \pi(i) \}~\text{for}~i = 1,2, \dots, d$.
  \end{algorithmic}
  \end{algorithm}
  Now, we show the correctness of the above procedure, We first claim that each $D_i$ ($i=0,1,2,\dots, d$) is full-dimensional.
Since $D_d = (\underline{z}+[0,1]^d) \cap \overline{\mathcal{K}}$, $(\underline{z}+[0,1]^d) \cap \overline{\mathcal{K}}$ is full-dimensional as desired.
To show this claim, we need the following auxiliary lemma.
For $c \in \mathbb{R}^d$ and $r>0$, let $B(c,r) \subseteq\mathbb{R}^d$ be the open ball with center $c$ and radius $r$, i.e., $B(c,r) = \{ y \in \mathbb{R}^d \mid ||y-c||_2 < r \}$.
Note that a subset of $\mathbb{R}^d$ is full-dimensional if and only if it includes an open ball (which in turn is equivalent to that it has a positive volume).

\begin{lemma}
  \label{lem:full-dim-conv-cap-ball-positive-volume}
  Let $P \subseteq \mathbb{R}^d$ be a full-dimensional convex set.
  Then for any $x \in P$ and any $\varepsilon >0$, 
  ${\rm vol}(P \cap B(x,\varepsilon)) > 0$.
  \end{lemma}
  \begin{proof}
  Since $P$ is full-dimensional, there exists $c \in P$ and $\varepsilon' > 0$ such that $B(c,\varepsilon') \subseteq \mathbb{R}^d$.
  For any $\lambda \in (0,1)$, we have $B_\lambda:=(1-\lambda) x + \lambda B(c,\varepsilon') \subseteq P$, since $P$ is convex.
  Moreover, for sufficiently small $\lambda>0$, we have $B_\lambda\subseteq B(x,\varepsilon)$.
  (Concretely, $\lambda = \varepsilon/(||c-x||_2 + ||\varepsilon'||_2)$ implies $B_\lambda\subseteq B(x,\varepsilon)$.)
  Hence, $B_\lambda \subseteq P \cap B(x,\varepsilon)$.
  As $B_\lambda$ has a positive volume, so does $P \cap B(x,\varepsilon)$.
  \qed
  \end{proof}

We are ready to prove that $D_i$ is full-dimensional for each $i=0,1,2,\dots, d$.
\begin{lemma}
  $D_i$ is full-dimensional for each $i=0,1,2,\dots, d$.
  \end{lemma}
  \begin{proof}
  We show this by induction on $i$.
  Since $D_0 = \overline{\mathcal{K}}$, the statement holds for $i=0$ by our assumption on $\dom (f)$.
  Let $i > 0$ and 
  assume that $D_{i-1}$ is full-dimensional.
  We show that $D_{i}$ is also full-dimensional in the following.
  
  When $x_i \notin \mathbb{Z}$, we know that $B(x,\varepsilon)$ is contained in $\{ y \in \mathbb{R}^d \mid z_i \le y_i \le z_i + 1 \}$ if $\varepsilon = \min(x-\lfloor x \rfloor, \lceil x \rceil)$.
  Hence, $D_i \cap B(x,\varepsilon) = D_{i-1} \cap B(x,\varepsilon)$.
  $D_{i-1} \cap B(x,\varepsilon)$ has a positive volume from Lemma~\ref{lem:full-dim-conv-cap-ball-positive-volume}.
  Hence, so does $D_i \cap B(x,\varepsilon)$ and thus $D_i$ is full-dimensional.
  
  When $x_i \in \mathbb{Z}$, we know that there exists $x',x'' \in D_{i-1}$ such that $x'_i = \underline{z}_i$ and $x''_i = \underline{z}_i+1$ by full-dimensionality of $D_{i-1}$ and definition of $\underline{z}_i$.
  As $D_{i-1}$ is convex, we have $y:=(x'+x'')/2 \in D_{i-1}$.
  Since $y_i = z_i + 1/2$, $D_i \cap B(y,1/2) = D_{i-1} \cap B(y,1/2)$.
  $D_{i-1} \cap B(y,1/2)$ has a positive volume from Lemma~\ref{lem:full-dim-conv-cap-ball-positive-volume}.
  Hence, so does $D_i \cap B(y,1/2)$ and thus $D_i$ is full-dimensional.\qed
  \end{proof}

  Since $(\underline{z}+[0,1]^d) \cap \overline{\mathcal{K}}$ is full-dimensional as shown above, the function $f_{\underline{z}} := f|_{\mathcal{K}\cap(\underline{z}+[0,1]^d)}$ is a submodular function over a simple distributive lattice.
Also, we can compute the poset representation of $\dom(f_{\underline{z}})(=K\cap(z+[0,1]^d))$ as follows.
$i \preceq j$ if $\underline{z}_i-\underline{z}_j = \delta_{i,j}:=\max_{y \in \overline{\mathcal{K}}} y_i-y_j$.
By topologically sorting this poset, we can compute a maximal chain in $(\underline{z}+\{0,1\}^d) \cap \mathcal{K}$ and thus compute the subgradient of $\hat{f}$ at $x$ by (\ref{eq_lovasz}) in Lemma~\ref{lem_lovasz}.
We summarize this as a proposition as follows.

\begin{proposition}
  Let $f:\mathcal{K}\rightarrow \mathbb{R}$ be an $\mathrm{L}^{\natural}$-convex function.
  Suppose $\mathcal{K}$ is full-dimensional and the expression~\eqref{eq:L-convex-linear-ineq-expression} of $\mathcal{K}$ is given.
  Then there exists a polynomial time algorithm that, given point $x \in \overline{\mathcal{K}}$, computes a subgradient of the convex extension $\hat{f}$ of $f$ at $x$.
  \end{proposition}

  \section{Upper Bound on Regret}
  \subsection{Full Information Setting}
  In this section, we extend the online submodular minimization algorithm (Algorithm 2 from Hazan and Kale~\cite{hazan2012online}) to $\mathrm{L}^{\natural}$-convex functions in the full information setting.
  Subsequently, we derive an upper bound for the regret of the algorithm.

  Let $\mathcal{K}$ be an arbitrary  $\mathrm{L}^{\natural}$-convex set, and let $\overline{\mathcal{K}}$ be the convex hull of $\mathcal{K}$.
  As a preparation, define the convex projection $\Pi_{\overline{\mathcal{K}}}:\mathbb{R}^d\rightarrow \overline{\mathcal{K}}$ onto the convex set $\overline{\mathcal{K}}$ as follows:
  \begin{equation}
    \Pi_{\overline{\mathcal{K}}}(y) := \underset{x\in \overline{\mathcal{K}}}{\text{argmin}}\|x-y\|.
  \end{equation}
  When $\mathcal{K}=[N]^d$, $\Pi_{\overline{\mathcal{K}}}(y)$ can be easily calculated as follows:
  \begin{equation}
    \Pi_{\overline{\mathcal{K}}}(y)(i)=\left\{
      \begin{array}{ll}
      N~&\text{if}\ y(i) > N\\
      1~&\text{if}\ y(i) <1\\
      y(i)~&\text{otherwise}
      \end{array}
    \right.~(i=1, \dots, d).
  \end{equation}
  We assume that this projection operation can be done efficiently, since it is a minimization of a convex function.

  Next, define the rounding function $\mathcal{P}_\tau:\mathbb{R}^d \rightarrow \mathbb{Z}^d$ using the threshold $\tau\in[0, 1]$ as follows:
  \begin{equation}
  \mathcal{P}_\tau(x)(i) := \left\{
    \begin{array}{ll}
    \lceil x(i)\rceil~&if~x(i) > \lfloor x(i) \rfloor + \tau\\
    \lfloor x(i)\rfloor~&if~x(i) \leq \lfloor x(i) \rfloor + \tau
    \end{array}
  \right.~(i=1, \dots, d).
  \end{equation}
  With these preparatory definitions in place, we now detail the operational flow of the Algorithm~\ref{alg_full}~(L$^{\natural}$-convex Subgradient Descent).

  At each iteration $t$, a threshold $\tau$ is chosen uniformly at random from the interval $[0, 1]$.
  This threshold is then used to discretize the continuous decision variable $x_t\in \overline{\mathcal{K}}$ into a discrete variable $z_t\in \mathcal{K}$. 
  Subsequently, the cost associated with $z_t$ is determined. Based on this cost, $x_t$ is updated using the calculated subgradient. 
The algorithm repeats this process $T$ times.
The process flow can be summarized as shown in Algorithm~\ref{alg_full} below.
  \begin{algorithm}[htb]
    \caption{L$^{\natural}$-convex Subgradient Descent}
    \label{alg_full}
    \begin{algorithmic}[1]
      \Require{parameter $\eta>0$. Let $x_1\in\overline{\mathcal{K}}$ be an arbitary initial point.}
      \For{$t=1~\text{to}~T$}
      \State Choose a threshold $\tau\in[0, 1]$ uniformly at random.
      \State Set $z_t= \mathcal{P}_\tau(x_t)$.
      \State Obtain cost $f_t(z_t)$.
      \State Find a maximal chain associated with $x_t$, $\emptyset = A_0 \subsetneq A_1\subsetneq \dots\subsetneq A_d = [d]$ and $\underline{z}$ by Algorithm 1.
      \State Compute a subgradient $g_t$ of the convex extension $\hat{f}_t$ at $x_t$  using $f_t(\underline{z} +\chi_{A_0}), f_t(\underline{z} +\chi_{A_1}),\dots, f_t(\underline{z} +\chi_{A_d})$.
      \State Update: set $x_{t+1}=\Pi_{\overline{\mathcal{K}}}(x_t-\eta g_t).$
      \EndFor
    \end{algorithmic}
  \end{algorithm}
  Here, at each round $t$, the most computationally expensive part is the element-by-element sorting required to find the Lov\'asz extension, with a computational complexity of $O(d\log d)$.
Therefore, the overall computational complexity is $O(Td\log d)$, which can be computed in polynomial time.
  
Next, in preparation for showing the regret upper bound of Algorithm~\ref{alg_full}, we introduce the Lemma~\ref{lemZhangEx3}.
  Lemma~\ref{lemZhangEx3} states that the norm of the subgradient is bounded above by the Lipschitz constant.
  \begin{lemma}[Supplementary material of \cite{zhang2021stochastic} Example~3]\label{lemZhangEx3}
    Let $f$ be $\mathrm{L}^{\natural}$-convex function, and has $l_\infty$-Lipschitz constant $\hat{L}$.
    Let $\hat{f}:\overline{\mathcal{K}}\rightarrow \mathbb{R}$ be the convex extension of $f$.
    For any $x\in \overline{\mathcal{K}}$, subgradient $g$ of $\hat f$ at $x$ computed using Algorithm 1 satisfies $ \|g\|_1 \leq \frac{3}{2}\hat{L}.$
  \end{lemma}

  For regret analysis, we extend Lemma 11 in Hazan and Kale~\cite{hazan2012online} over $\mathcal{K}$.
  \begin{lemma}\label{lem_regret_full}
    Let $f_t:\mathcal{K}\rightarrow \mathbb{R}~(t=1,2, \dots , T)$ be a sequence of $\mathrm{L}^{\natural}$-convex functions.
    Let $\hat{f}_t:\overline{\mathcal{K}}\rightarrow \mathbb{R}~(t=1,2, \dots, T)$ be the convex extension of $f_t$.
    Let $x_t\in \overline{\mathcal{K}}~(t=1,2, \dots, T)$ be defined by $x_1=0$ and $x_{t+1} = \Pi_{\overline{\mathcal{K}}}(x_t-\eta \hat{g}_t)$, where $\hat{g}_1, \hat{g}_2, \dots, \hat{g}_T$ are vector valued random variables such that $\mathbf{E}[\hat{g}_t\,|\,x_t] = g_t$, where $g_t$ is a subgradient of $\hat{f}_t$ at $x_t$.
    Then the expected regret of playing $x_1, x_2, \dots, x_T$ is bounded as 
    \begin{equation}
      \sum_{t=1}^{T}\mathbf{E}[\hat{f}_t(x_t)]-\min_{x\in\mathcal{K}}\sum_{t=1}^{T}\hat{f}_t(x)\leq \frac{dN^2}{2\eta}+\frac{\eta}{2}\sum_{t=1}^T\mathbf{E}[\|g_t\|^2].
    \end{equation}
  \end{lemma}
  The proof is similar to Lemma 11 in Hazan and Kale~\cite{hazan2012online}.
  \begin{proof}
    Let $y_{t+1} = x_t-\eta \hat{g}_t$, so that $x_{t+1}=\Pi_{\mathcal{K}}(y_{t+1})$. Note that
  \begin{equation}\label{eq_lem_regret_full_1}
    \|y_{t+1}-x^*\|^2 =\|x_t-x^*\|^2 -2\eta \hat{g}^\top(x_t-x^*)+\eta^2\|\hat{g}_t\|^2.
  \end{equation}
Rearranging Equation~(\ref{eq_lem_regret_full_1}), we have
  \begin{equation}
  \hat{g}_t^\top(x_t-x^*) = \frac{1}{2\eta}(\|x_t-x^*\|^2 -\|y_{t+1}-x^*\|^2)+\frac{\eta}{2}\|\hat{g}_t\|^2.
  \end{equation}
  Utilizing the property that $\|x_{t+1}-x^*\| \leq \|y_{t+1}-x^*\|$ (a consequence of the properties of Euclidean projections onto convex sets) leads to
  \begin{equation}
  \hat{g}_t^\top(x_t-x^*) \leq \frac{1}{2\eta}(\|x_t-x^*\|^2 -\|x_{t+1}-x^*\|^2)+\frac{\eta}{2}\|\hat{g}_t\|^2.
  \end{equation}
  Aggregating the terms for $t=1,2, \dots,T$, we have
  \begin{align}
    \sum_{t=1}^{T} \hat{g}^\top(x_t-x^*) &\leq \sum_{t=1}^{T} \frac{1}{2\eta} (\|x_t-x^*\|_2^2-\|x_{t+1}-x^*\|_2^2)+\frac{\eta}{2}\|\hat{g}_t\|_2^2\\
    &= \frac{1}{2\eta} (\|x_1-x^*\|_2^2-\|x_{T}-x^*\|_2^2)+\sum_{t=1}^{T}\frac{\eta}{2}\|\hat{g}_t\|_2^2\\
    &\leq  \frac{1}{2\eta} (\|x_1-x^*\|_{2}^2)+\sum_{t=1}^{T}\frac{\eta}{2}\|\hat{g}_t\|_2^2\label{eq_lem_regret_full_2}\\
    &\leq  \frac{1}{2\eta} (d\|x_1-x^*\|_{\infty}^2)+\sum_{t=1}^{T}\frac{\eta}{2}\|\hat{g}_t\|_2^2\label{eq_lem_regret_full_3}\\
    &\leq  \frac{dN^2}{2\eta} +\sum_{t=1}^{T}\frac{\eta}{2}\|\hat{g}_t\|_2^2,
  \end{align}
  where inequality~(\ref{eq_lem_regret_full_2}) is derived from the relationship $\|x\|_2 \leq d\|x\|_{\infty}$, and inequality~(\ref{eq_lem_regret_full_3}) leverages the bound $\|x-y\|_{\infty}^2 \leq N^2$ from the definition of $N$.
  Next, given that $\mathbf{E}[\hat{g}_t|x_t]=g_t$ (a subgradient of $\hat{f}_t$ at $x_t$), we obtain
  \begin{equation}
    \mathbf{E}[\hat{g}_t^\top(x_t-x^*)|x_t]= g_t^\top(x_t-x^*) \geq \hat{f}_t(x_t)-\hat{f}_t(x^*),
  \end{equation}
  owing to the convexity of $\hat{f}_t$.
  By taking the expectation over the selection of $x_t$, we derive
  \begin{equation}
    \mathbf{E}[\hat{g}_t^\top(x_t-x^*)] \geq \mathbf{E}[\hat{f}_t(x_t)]-\hat{f}_t(x^*).
  \end{equation}
  Consequently, the expected regret can be bounded as follows:
  \begin{equation}
    \sum_{t=1}^T \mathbf{E}[\hat{f}_t(x_t)] - \hat{f}_t(x^*)\leq \mathbf{E}\left[\sum_{t=1}^T \hat{g}_t^\top(x_t - x^*)\right]\leq \frac{dN^2}{2\eta} +\frac{\eta}{2}\sum_{t=1}^{T}\mathbf{E}[\|\hat{g}_t\|_2^2].
  \end{equation}
  \qed
  \end{proof}

  Building on the aforementioned results, we give the regret bounds in Theorem~\ref{theorem_alg1}.
  \begin{theorem}\label{theorem_alg1}
  When Algorithm~\ref{alg_full} is executed with the parameter $\eta=\sqrt{\frac{dN^2}{T(\frac{3}{2}\hat{L})^2}}$, it achieves a regret bound of $\mathbf{E}[R_T]\leq \frac{3}{4}N\hat{L}\sqrt{dT}.$
  \end{theorem}
  It can be proved by using Lemmas \ref{lem_lovasz},~\ref{lemZhangEx3}, and \ref{lem_regret_full}.
  \begin{proof}
    Using Lemmas~\ref{lem_lovasz},~\ref{lemZhangEx3}, and \ref{lem_regret_full}, we derive
    \begin{align}
      \mathbf{E}[R_T] &=  \sum_{t=1}^T \mathbf{E}[f_t (z_t)]-\min_{z^*\in \mathcal{K}}\sum_{t=1}^T f_t(z^*)\\
      &= \sum_{t=1}^T \hat{f}_t (x_t)-\min_{x^*\in \overline{\mathcal{K}}}\sum_{t=1}^T \hat{f}_t(x^*)\label{eq_the1_1}\\
      &\leq  \frac{dN^2}{2\eta} +\frac{\eta}{2}\sum_{t=1}^{T}\mathbf{E}[\|\hat{g}_t\|_2^2]\label{eq_the1_2}\\
      & \leq  \frac{dN^2}{2\eta} + \frac{\eta}{2} T (\frac{3}{2}\hat{L})^2\label{eq_the1_3}= \frac{3}{4}N\hat{L}\sqrt{dT},
    \end{align}
    where the transition to equation~(\ref{eq_the1_1}) is justified by the equivalence $\mathbf{E}[f_t(z_t)]=\hat{f_t}(x_t)$ as established in Lemma~\ref{lem_lovasz}. 
    The bound in inequality~(\ref{eq_the1_2}) is obtained from Lemma~\ref{lem_regret_full}, and the final inequality~(\ref{eq_the1_3}) is supported by the norm condition $\|\hat{g}_t\|_2 \leq \|\hat{g}_t\|_1 \leq \frac{3}{2}\hat{L}$, as delineated in Lemma~\ref{lemZhangEx3}.
    \qed
  \end{proof}
  \subsection{Bandit Setting}
  In this section, we extend the online submodular minimization algorithm (Algorithm 3 in Hazan and Kale~\cite{hazan2012online}) to L$^{\natural}$-convex functions to obtain upper bound on regret.
  Let $f_t:\mathcal{K}\rightarrow [-M, M]~(\forall t \in[T])$, where the function value is bounded.

  We describe the subgradient descent algorithm under the bandit setting for $\mathrm{L}^{\natural}$-convex functions.
  For each iteration $t$, the algorithm identifies the maximal chain associated with $x_t$ and its associated permutation $\pi$.
  This allows for the representation of $x_t$ as a convex combination.
  A point $z_t$ is chosen based on probabilities $\rho_i$, which are derived from coefficients $\mu_i$ in the representation of $x_t$ and parameter $\delta$.
  The cost $f_t(z_t)$ at the chosen point $z_t$ is obtained.
An unbiased estimator of the subgradient $\hat{g}_t$ of $\hat{f}_t$ at $x_t$ is computed, varying according to the value of $z_t$, probabilities $\rho_i$, and a randomly selected $\varepsilon_t$.
The algorithm updates $x_{t+1}$ using the current point $x_t$, step size $\eta$, and the estimated subgradient $\hat{g}_t$.
The process flow can be summarized as shown in Algorithm~\ref{alg_bandit} below.  
  \begin{algorithm}[htb]
    \caption{Bandit L$^{\natural}$-convex Subgradient Descent}\label{alg_bandit}
    \begin{algorithmic}[1]
      \Require{parameters $\eta,\delta>0$. Let $x_1\in\overline{\mathcal{K}}$ be arbitary initial point.}
      \For{$t=1~\text{to}~T$}
      \State Find a maximal chain associated with $x_t$, $\emptyset = A_0 \subsetneq A_1\subsetneq \dots\subsetneq A_d = [d]$ and $\underline{z}$ by Algorithm 1.
      \State Choose the point $z_t$ as follows:
      $$z_t = \underline{z} +\chi_{A_i}  \text{with\, probability}\quad \rho_i = (1-\delta)\mu_i+\frac{\delta}{d+1}.$$
      \State Obtain cost $f_t(z_t)$.
      \State Compute a unbiased estimator of a subgradient $\hat{g}_t$ of $\hat{f}_t$ at $x_t$ as follows.
      If $z_t =  \underline{z} +\chi_{A_0}$, then set $\hat{g}_t = -\frac{1}{\rho_0}f_t(z_t){e_{\pi(1)}}$, and if 
      $ z_t = \underline{z} +\chi_{A_n}$, then set $\hat{g}_t=\frac{1}{\rho_n}f_t(z_t){e_{\pi(n)}}$. Otherwise, $z_t= \underline{z} +\chi_{A_i}$ for same $1\leq i\leq n-1$. Choose $\varepsilon_t \in \{-1,1\}$ uniformly at random, and set:
      $$\hat{g}_t  =  \left\{
        \begin{array}{ll}
        \frac{2}{\rho_i}f_t(z_t){e_{\pi(i)}} &\text{if}\quad \varepsilon_t =1 \\
        -\frac{2}{\rho_i}f_t(z_t){e_{\pi(i+1)}}& \text{if} \quad \varepsilon_t = -1. 
        \end{array}
        \right.$$
      \State Update: set $x_{t+1}=\Pi_{\overline{\mathcal{K}}}(x_t-\eta \hat{g}_t)$.
      \EndFor
    \end{algorithmic}
  \end{algorithm}
  On Algorithm~\ref{alg_bandit}, replacing the loss function $f_t$ with its convex extension $\hat{f}_t$, the error is bounded by $2\delta M$.
  \begin{lemma}\label{lem_regret_bandit}
  For all $t\in [T]$, we have $ \mathbf{E}[f_t(z_t)]\leq \mathbf{E}[\hat{f}_t(x_t)]+2\delta M$.
  \end{lemma}
  The proof is similar to Lemma 15 in \cite{hazan2012online}.
  \begin{proof}
    Let $\emptyset =A_0 \subsetneq A_1 \subsetneq \dots \subsetneq A_d = [d], \underline{z}\in\mathcal{K}$ is a maximal chain associated with $x$.
  From Definition~\ref{def_convexextension_onL}, we derive that $\hat {f}_t (x_t)=\sum_{i} \mu_i f_t(\underline{z} +\chi_{A_i})$.
  On the other hand, $\mathbf{E}_t[f_t(z_t)]=\sum_i \rho_if_t(\underline{z} +\chi_{A_i})$, and hence:
  \begin{align}
    \mathbf{E}_t[f_t(z_t)]-\hat{f}_t(x_t) &= \sum_{i=0}^d (\rho_i -\mu_i)f_t(\underline{z} +\chi_{A_i})\\
    &=\delta \sum_{i=0}^d \left(\frac{1}{d+1} +\mu_i\right)f_t(\underline{z} +\chi_{A_i})\\
    &\leq \delta \sum_{i=0}^d \left(\frac{1}{d+1} +\mu_i\right)|f_t(\underline{z} +\chi_{A_i})|\\
    &\leq 2\delta M.
  \end{align}
  Then, taking the expected value for each filtration $\mathcal{F}_{t-1}$, we obtain $\mathbf{E}[f_t(z_t)]-\mathbf{E}[\hat{f}_t(x_t)] \leq 2\delta M$.
  \qed
  \end{proof}
  
  For the regret analysis, we show that the unbiased estimator of the subgradient can be suppressed from above as follows.
  \begin{lemma}\label{lem_subgradient}
    For all $t$, we have $\mathbf{E}[\|\hat{g}_t\|^2]\leq \frac{16M^2d^2}{\delta}$.
  \end{lemma}
  \begin{proof}
    Let $\emptyset =A_0 \subsetneq A_1 \subsetneq \dots \subsetneq A_d = [d], \underline{z}\in\mathcal{K}$ is a maximal chain associated with $x$.
    Since $\hat{g}_t$ is an unbiased estimator of $g_t$, we have $\mathbf{E}[\hat{g}_t|x_t]=\mathbf{E}_t[\hat{g}_t]=g_t$.
    Thus, the following is obtained:
    \begin{equation}
      \mathbf{E}_t[\|\hat{g}_t\|^2]\leq \sum_{i=0}^d \frac{4}{\rho_i^2}f_t(\underline{z} +\chi_{A_i})^2\cdot \rho_i\leq \frac{4M^2(d+1)^2}{\delta}\leq \frac{16M^2d^2}{\delta}.
    \end{equation}
    Then, as in Lemma \ref{lem_regret_bandit}, taking the expected value for each filtration $\mathcal{F}_{t-1}$, we obtain the desired inequality $\mathbf{E}[\|\hat{g}_t\|^2]\leq \frac{16M^2d^2}{\delta}$.
    \qed
  \end{proof}

  Thus, the following regret upper boundary is obtained.
  \begin{theorem}\label{theorem_alg2}
    Algorithm \ref{alg_bandit}, run with parameters $\delta = \frac{d}{T^{1/3}},\eta = \frac{N}{4MT^{2/3}}$, achieves the following regret bound : $ \mathbf{E}[R_T] \leq 6dNMT^{2/3}$.
  \end{theorem}
  It can be proved by using Lemmas \ref{lem_regret_full}, \ref{lem_regret_bandit} and \ref{lem_subgradient}.
  \begin{proof}
    Using Lemmas \ref{lem_regret_full}, \ref{lem_regret_bandit} and \ref{lem_subgradient}, we have
  \begin{align}
    \mathbf{E}[R_T] &= \mathbf{E}[f_t(z_t)]- \min_{z^*\in \mathcal{K}} \sum_{t=1}^{T}f_t(z^*)\\
  &\leq 2\delta MT +\sum_{t=1}^{T}E[\hat{f}_t(x_t)]- \min_{x^*\in \overline{\mathcal{K}}} \sum_{t=1}^{T}\hat{f}_t(x^*)\label{eq_theorem_alg2_1}\\
    &\leq 2\delta MT + \frac{dN^2}{2\eta}+\frac{\eta}{2}\sum_{t=1}^T\mathbf{E}[\|\hat{g}_t\|^2] \label{eq_theorem_alg2_2}\\
    &\leq 2\delta MT + \frac{dN^2}{2\eta}+\frac{8d^2M^2\eta T}{\delta}\label{eq_theorem_alg2_3}\\
    &=2dMT^{2/3}\cdot(2N+1)\\
    &\leq 6dNMT^{2/3},
  \end{align}
  where the bound in inequality~(\ref{eq_theorem_alg2_1})  is obtained from Lemma~\ref{lem_regret_bandit}.
  The bound in inequality~(\ref{eq_theorem_alg2_2}) is obtained from Lemma~\ref{lem_regret_full}.
  The bound in inequality~(\ref{eq_theorem_alg2_3}) is obtained from Lemma~\ref{lem_subgradient}.
  \qed
  \end{proof}
\section{Lower Bound on Regret}
We provide a lower bound for regret, indicating that any algorithms designed for online $\mathrm{L}^{\natural}$-convex minimization necessarily incur a minimum regret of $\Omega(\hat{L}N\sqrt{dT})$. 
This result implies that the upper bound presented in Theorem~\ref{theorem_alg1} is optimal up to a constant factor.
\begin{theorem}
  For any algorithm solving online $\mathrm{L}^{\natural}$-convex minimization, there exists a sequence of $\mathrm{L}^{\natural}$-convex functions $f_1, f_2, \dots, f_T\colon [N]^d \rightarrow \mathbb{R}$, with an $l_\infty$-Lipschitz constant $\hat{L}$, such that the regret is at least $\Omega( \hat{L}N\sqrt{dT})$.
\end{theorem}
The proof is similar to Theorem 14 in Hazan and Kale~\cite{hazan2012online}.
\begin{proof}
  Consider a random sequence of cost functions.
  At each iteration $t\in [1, T]$, select $i(t) = (t\bmod d) + 1 \in [d]$ and a Rademacher random variable $\sigma_t \in \{-1, 1\}$, chosen independently from all other random variables. Define $f_t\colon [N]^d \rightarrow \mathbb{R}$ as
  \begin{equation}
    f_t(s) = \sigma_t Ls_{i(t)}\ \ \ (s\in[N]^d),
  \end{equation}
  where $s_i$ denotes the $i$-th element of $s$ for $i \in [d]$.

  Since $f_t$ is a linear function, it is an $\mathrm{L}^{\natural}$-convex function as well.
  Furthermore, due to the properties of Rademacher random variables, it holds that $\mathbf{E}[f_t(s)]=0$.
  Consequently, the following is true for regret:
  \begin{equation}
    \mathbf{E}[R_T]=\mathbf{E}\left[ \sum_{t=1}^{T}f_t(s_t)\right] -\min_{s\in[N]^d}\sum_{t=1}^{T}f_t(s)
  \\=-\min_{s\in[N]^d}\sum_{t=1}^{T}f_t(s).
  \end{equation}

  To compute the regret, we construct $s^*=\argmin_{s\in[N]^d}\sum_{t=1}^{T}f_t(s)$ as follows:
  \begin{equation}
  X_i = \sum_{t:i(t)=i}^{T} \sigma_t,\quad s_{i}^* =  \left\{
    \begin{array}{ll}
    1\ &\mbox{if}\ X_i \geq 0 \\
    N\ &\mbox{if}\ X_i < 0.
    \end{array}
  \right.
  \end{equation}

When the slope along the $i$-axis of $\sum_{t=1}^{T}f_t(s)$ is positive, $1$ becomes the minimal value, and when it is negative, $N$ becomes the minimal value.
Thus, $s^*=\underset{s\in[N]^d}{\text{argmin}}\sum_{t=1}^{T}f_t(s)$ holds.
Therefore, we have
  \begin{align}
  \mathbf{E} \left[ \sum_{t=1}^{T}f_t(s^*)\right] =~&\mathbf{E}\left[ \sum_{t=1}^{T} \sigma_t \hat{L}S_{i}^*(t)\right] =\mathbf{E}\left[ \sum_{i=1}^{d}\sum_{t:i(t)=i}^{T}\sigma_t\hat{L}S_{i}^*\right] =~\mathbf{E}\left[ \sum_{i=1}^{d}X_i\hat{L}S_{i}^*\right] \\
  =~&\mathbf{E}\left[ \hat{L}\sum_{i=1}^{d}X_i\left( 1\cdot\1[X_i\geq 0]+N\cdot\1[X_i<0]\right) \right] \\
  =~&\mathbf{E}\left[ \hat{L}\sum_{i=1}^{d}-\frac{N-1}{2}|X_i|+\frac{N+1}{2}X_i\right] \\
  =~&-\Omega \left( \hat{L}\cdot d \cdot \left( \frac{N-1}{2}\right) \cdot \left(\sqrt{\frac{T}{d}}\right) \right) \label{eq_the_low_1}\\
  =~&-\Omega \left( \hat{L}N\sqrt{dT}\right),
  \end{align}
  where the inequality~(\ref{eq_the_low_1}) is derived from Khintchine's inequality (see, e.g., \cite{cesa2006prediction}) and $\mathbf{E}[X_i]=0$.
  \qed
\end{proof}
  \section{Applications}\label{application}
  This section presents several applications that can be captured in the framework of online $\mathrm{L}^{\natural}$-convex minimization.

  One is an extension to a natural online version of the spare parts inventory control problem.
  A straightforward online version of the existing model would require feedback of expected values.
  Requiring feedback of expected values weakens the advantage of going online, which does not require assumptions about the demand distribution.
  Our proposed spare-parts inventory management problem does not require expectation feedback.

  The second is an application of queueing theory to the call center shift scheduling problem.
  In queueing theory, multimodular functions often appear, which are equivalent to L$^{\natural}$-convex functions by a simple linear transformation (unimodular transformation).
  As one such example, we confirm that the shift scheduling problem in a call center falls within the framework of online L$^{\natural}$-convex minimization.

  \subsection{Online Inventory System of Reparable Spare Parts}
  The spare parts inventory management problem, proposed by Miller~\cite{miller1971minimizing}, is used for parts management in aircraft maintenance, where the quantity demanded and the quantity ordered take discrete values.
  Miller's model seeks to minimize the cost of manufacturing a product with $d\in \mathbb{Z}_+$ parts, which is formulated as the sum of a fine determined by the maximum number of shortages of each part and the cost of purchasing spare parts in advance.
  Let $c_j>0$ be the unit price of variety $j\in [d]$ and $z_j \in \mathbb{Z}_+ \leq N$ be the quantity of spare parts ordered for variety $j$~($N\in \mathbb{Z}_+$ is the maximum amount that can be purchased.).
  The cost of purchasing spare parts is $\sum_{j=1}^d c_jz_j$.
  On the other hand, let the probability that the demand for variety $j$ is $m$ be $\varphi_j(m)$ and let the cumulative distribution function be $F_j(k)=\sum_{m=0}^k \varphi_j(m)$, then the expected maximum number of shortages for each part is $\sum_{k=0}^\infty (1-\prod_{j=1}^n F_j(x_j+k))$.
  Therefore, the objective of the offline spare parts inventory control problem is to solve the following problem:
  \begin{equation}\label{inventory_offline_problem}
    \min_{z\in [N]^d}\sum_{k=0}^\infty (1-\prod_{j=1}^d F_j(z_j+k)) + \sum_{j=1}^d c_jz_j.
  \end{equation}
  It is known that this problem can be formulated in the framework of  L$^{\natural}$-convex minimization~\cite{moriguchi2005discrete}.

Consider equation \eqref{inventory_offline_problem} as an online problem.
 After the decision is made, the sum of the purchase cost of each component and the expected maximum number of shortages of each component is obtained as feedback.
 
 However, the expected value of the maximum number of shortages of each component is not appropriate as feedback.
If it can be observed as feedback, there is no need to solve it as an online problem since the expected value is the average after sufficient iterations.

  Therefore, we redefine the problem as a more natural online decision-making problem and show that it can be captured in an online L$^{\natural}$-convex minimization framework.
  Existing formulations consider maximizing expected profits with a known demand distribution.
  The expected value is given as feedback, but as a practical matter, it is inconvenient for the user to feed back the expected value.
  On the other hand, we do not assume a distribution, which means that we can adapt to any environment.

  This model is an inventory model that minimizes the regret for long-term losses determined by the number of orders placed in each round.
  The problem is formulated as an online L$^{\natural}$-convex minimization problem.
  Each parameter is listed in Table \ref{table_inventory_parameter}.
    \begin{table}[htb]
      \caption{Parameters of the Online Inventory System for Repairable Spare Parts}
      \label{table_inventory_parameter}
      \centering 
      \begin{tabular}{lll}
        \toprule
        Parameter & Meaning & Variable Type \\
        \midrule
        $d \in \mathbb{Z}$ & Number of parts & Given \\
        $c \in \mathbb{R}^d$ & Unit cost of parts & Given \\
        $p \in \mathbb{R}$ & Penalty cost & Given \\
        $z \in \mathbb{Z}^d$ & Order quantity of parts & Decision \\
        $f_t(z):[N]^d \rightarrow \mathbb{R}, (t \in [T])$ & Cost function on round $t$ & Feedback of round $t$ \\
        $y_t \in \mathbb{Z}^d, (t \in [T])$ & Demand of parts on round $t$ & Feedback of round $t$ \\
        \bottomrule
      \end{tabular}
    \end{table}
    The loss function consists of the sum of the penalty cost determined by the maximum number of missing parts and the cost of purchasing spare parts, and is formulated as follows:
\begin{equation}
  f_t(z)=  p( \max_{j\in[d]}\max(y_{t, j}-z_j, 0))+\sum_{j=1}^d c_jz_j.
\end{equation}

  In this case, $f_t$ is an $\mathrm{L}^{\natural}$-convex function, so the problem is online $\mathrm{L}^{\natural}$-convex minimization. 
  Show that the loss function $f_t(x)$ is an L$^{\natural}$-convex function.
  In preparation for the proof, we introduce the following lemma.

  \begin{lemma}[Maxmum-component function \cite{murota1998discrete}]\label{lem_murota1998}
    Let $p=(p_1, p_2,\dots,p_n)\in \mathbb{Z}^n$, and let $g:\mathbb{Z}^n\rightarrow \mathbb{R}$.
    For any  $\tau_0,\tau_1, \dots,\tau_n \in \mathbb{R}$,
    $$g(p)=\max(\{\tau_0,p_1+\tau_1,\dots,p_n+\tau_n\})$$
    is an L$^{\natural}$-convex function.
    \end{lemma}
    Next, show the follwing lemma.
    \begin{lemma}\label{lem_f(-x)_Lconvex}
    Let $g:\mathbb{Z}^n\rightarrow \mathbb{R}$, and let $f:\mathbb{Z}^n\rightarrow \mathbb{R}$ is an L$^{\natural}$-convex function.
    $g(x)=f(-x)\Rightarrow g(x)$ is an L$^{\natural}$-convex function.
    \end{lemma}
    \begin{proof}
    We show that $g(x)$ satisfies discrete midpoint convexity:
    \begin{align*}
      g(x)+g(y)&=f(-x)+f(-y)\\
      &\geq f\left(\left\lfloor\frac{-(x+y)}{2}\right\rfloor\right)+f\left(\left\lceil\frac{-(x+y)}{2}\right\rceil\right)\\
      &=f\left(-\left\lceil\frac{-(x+y)}{2}\right\rceil\right)+f\left(-\left\lfloor\frac{-(x+y)}{2}\right\rfloor\right)\\
      &=g\left(\left\lceil\frac{x+y}{2}\right\rceil\right) + g\left(\left\lfloor\frac{x+y}{2}\right\rfloor\right).
    \end{align*}
    \qed
    \end{proof}
    We show that the loss function is an L$^{\natural}$-convex function by using Lemma \ref{lem_murota1998} and Lemma \ref{lem_f(-x)_Lconvex}.
    \begin{lemma}
    $$f_t(x)=  p( \max_{j\in[n]}\max(y_j-x_j, 0))+\sum_{j=1}^n c_jx_j$$
      is an L$^{\natural}$-convex function.
    \end{lemma}
    \begin{proof}
      By using Lemma \ref{lem_murota1998} and Lemma \ref{lem_f(-x)_Lconvex}, $\max_{j\in[n]}\max(y_j-x_j, 0)$ is an L$^{\natural}$-convex function. 
      $\sum_{j=1}^n c_jx_j$ is an L$^{\natural}$-convex function.
    Thus, $f_t(x)$ is an L$^{\natural}$-convex function.\qed
    \end{proof}

    \subsection{Shift Scheduling with a Global Service Level Constraint}
    Here, we introduce the shift scheduling with a global service level constraint proposed by Koole and van der Sluis~\cite{Koole2003optimal} from queueing theory and its online implementation.

    Consider one service center.
    This model aims to minimize the total loss and labor costs due to overall service level degradation over time.
    A call center operates over $I$ time intervals. Denote the time intervals by $i=1, \dots, I$. 
    Each operator works for $M$ consecutive time intervals.
    There are $K$ types of work shifts, and the starting point of each is specified in advance. Shifts are denoted by $k=1,\dots,K$.  and the starting time interval of a $K$ shift is denoted by $I_k$.
    For each iteration $t$ and for each time interval $i$, there is a function $g_{t, i}(n_{i})$ that represents the service level in that interval. where $n_{i}$ is the number of operators working in time interval $i$. The function $g_{i}$ is a monotonically increasing concave function.
    For each iteration $t$, the overall service level $S_t$ is given by $S_t=\sum_{1\leq i\leq I}g_{t, i}(n_i)$.
    Each parameter is listed in Table \ref{table_scheduling_parameter}.
    \begin{table}[htb]
      \caption{Parameters for Shift Scheduling with a Global Service Level Constraint}
      \label{table_scheduling_parameter}
      \centering  
      \begin{tabular}{lll}
        \toprule
        Parameter & Meaning & Variable Type \\
        \midrule
        $N \in \mathbb{Z}_+$ & Number of operators & Given \\
        $K \in \mathbb{Z}_+$ & Number of shift types & Given \\
        $I \in \mathbb{Z}_+$ & Duration interval & Given \\
        $l \in \mathbb{Z}^K$ & Labor costs for shifts & Given \\
        $y \in [N]^K$ & Number of people assigned to shifts & Decision \\
        $c \in \mathbb{Z}$ & Limit time to keep customers waiting & Given \\
        $G \in \mathbb{Z}$ & Projected profit if customers are satisfied & Given \\
        $r \in [0, 1]$ & \begin{tabular}{l}Probability of not making a profit \\ if the customer is not satisfied\end{tabular} & Given \\
        \bottomrule
      \end{tabular}
    \end{table}
    Let $y_k$ be the number of operators to be placed in the $k$-th shift, then the number of operators in time interval $i$, $h_i$ is $h_i(y)=\sum_{k:i-M<I_k\leq i}y_k$.
    Thus, the service level $S$ is equal to
    \begin{equation}
      S_t(y)=\sum_{1\leq i\leq I} g_{t, i}(h_i(y)).
    \end{equation}
    Let $\lambda$ be the arrival rate of customers for each time interval and $\mu$ be the service rate of the operator.
Furthermore, by setting a threshold $c$ for the limit of time to keep a customer waiting, the percentage of customers who connect to an operator within this time can be defined as a service level based on the queueing model $M/M/n$.
Therefore, the objective function $f'_t$ to be minimized is as follows:
    \begin{equation}
      f'_t(y) := -GrS_t(y) + \sum_{1\leq k \leq K}l_ky_k.
    \end{equation}
    This objective function is known to be a multimodular function~\cite{Koole2003optimal}.
    Here, the multimodular function $f'_t$ becomes $\mathrm{L}^{\natural}$-convex $f_t$ by unimodular transformation and can be solved in the framework of an online $\mathrm{L}^{\natural}$-convex minimization problem.
    The online decision-making problem of this model is that the allocation $y$ to a shift can be determined to minimize the long-term loss without a priori knowledge of the function $g_i$ representing the service level in the time interval $i$.

    \section{Conclusions}
    We proposed computationally efficient algorithms for online $\mathrm{L}^{\natural}$-convex minimization, which extends online submodular minimization.
    Our algorithms apply for two major settings: the full information setting and the bandit setting.
    We provided regret analyses of our algorithms and lower bound for the regrets, and in particular 
    showed that in the full information setting our proposed algorithm achieves a tight regret bound up to a constant factor. 
    We also demonstrated that the online $\mathrm{L}^{\natural}$-convex minimization naturally captures various problems, including
    the spare parts inventory management problem and the shift scheduling problem.
    \section*{Acknowledgement}
    This work was partially supported by the JST ERATO Grant Number JPMJER2301.
    Additionally, portions of this research were conducted during visits by the first author, Yokoyama, and the fourth author, Kimura, to NEC Corporation.
\bibliographystyle{splncs04}
\bibliography{reference}

\begin{thebibliography}{10}
\providecommand{\url}[1]{\texttt{#1}}
\providecommand{\urlprefix}{URL }
\providecommand{\doi}[1]{https://doi.org/#1}

\bibitem{altman2003discrete}
Altman, E.: Discrete-event Control of Stochastic Networks: Multimodularity and Regularity. Springer Science \& Business Media (2003)

\bibitem{altman2000multimodularity}
Altman, E., Gaujal, B., Hordijk, A.: Multimodularity, convexity, and optimization properties. Mathematics of Operations Research  \textbf{25}(2),  324--347 (2000)

\bibitem{bach2013learning}
Bach, F.: Learning with submodular functions: A convex optimization perspective. Foundations and Trends{\textregistered} in machine learning  \textbf{6}(2-3),  145--373 (2013)

\bibitem{bubeck2021kernel}
Bubeck, S., Eldan, R., Lee, Y.T.: Kernel-based methods for bandit convex optimization. Journal of the ACM  \textbf{68}(4),  1--35 (2021)

\bibitem{cesa2006prediction}
Cesa-Bianchi, N., Lugosi, G.: Prediction, Learning, and Games. Cambridge University Press (2006)

\bibitem{chen2018online}
Chen, L., Hassani, H., Karbasi, A.: Online continuous submodular maximization. In: International Conference on Artificial Intelligence and Statistics. pp. 1896--1905. PMLR (2018)

\bibitem{chen2017convexity}
Chen, X.: $\mathrm{L}^{\natural}$-convexity and its applications in operations. Frontiers of Engineering Management  \textbf{4}(3),  283--294 (2017)

\bibitem{chen2021discrete}
Chen, X., Li, M.: Discrete convex analysis and its applications in operations: A survey. Production and Operations Management  \textbf{30}(6),  1904--1926 (2021)

\bibitem{freund2017minimizing}
Freund, D., Henderson, S.G., Shmoys, D.B.: Minimizing multimodular functions and allocating capacity in bike-sharing systems. In: 19th International Conference on Integer Programming and Combinatorial Optimization. pp. 186--198. Springer (2017)

\bibitem{fujishige2005submodular}
Fujishige, S.: Submodular Functions and Optimization. Elsevier (2005)

\bibitem{fujishige2000notes}
Fujishige, S., Murota, K.: Notes on {L}-/{M}-convex functions and the separation theorems. Mathematical Programming  \textbf{88},  129--146 (2000)

\bibitem{hazan2012online}
Hazan, E., Kale, S.: Online submodular minimization. Journal of Machine Learning Research  \textbf{13}(10) (2012)

\bibitem{Koole2003optimal}
Koole, G., van~der Sluis, E.: Optimal shift scheduling with a global service level constraint. IIE Transactions  \textbf{35}(11),  1049--1055 (2003)

\bibitem{miller1971minimizing}
Miller, B.L.: On minimizing nonseparable functions defined on the integers with an inventory application. SIAM Journal on Applied Mathematics  \textbf{21}(1),  166--185 (1971)

\bibitem{moriguchi2005discrete}
Moriguchi, S., Murota, K.: Discrete {H}essian matrix for {L}-convex functions. IEICE transactions on fundamentals of electronics, communications and computer sciences  \textbf{88}(5),  1104--1108 (2005)

\bibitem{murota1998discrete}
Murota, K.: Discrete convex analysis. Mathematical Programming  \textbf{83},  313--371 (1998)

\bibitem{murota2003discrete}
Murota, K.: Discrete Convex Analysis. SIAM (2003)

\bibitem{murota2022discrete}
Murota, K.: Discrete convex analysis: A tool for economics and game theory. arXiv preprint arXiv:2212.03598  (2022), (Preliminary version: Murota, K.: Discrete convex analysis: A tool for economics and game theory. The Journal of Mechanism and Institution Design \textbf{1}(1), 151--273 (2016))

\bibitem{qin2014contextual}
Qin, L., Chen, S., Zhu, X.: Contextual combinatorial bandit and its application on diversified online recommendation. In: Proceedings of the 2014 SIAM International Conference on Data Mining. pp. 461--469. SIAM (2014)

\bibitem{sakaue2023rethinking}
Sakaue, S., Oki, T.: Rethinking warm-starts with predictions: Learning predictions close to sets of optimal solutions for faster $\mathrm{L}$-/$\mathrm{L}^{\natural}$-convex function minimization. In: International Conference on Machine Learning. pp. 29760--29776. PMLR (2023)

\bibitem{shalev2012online}
Shalev-Shwartz, S.: Online learning and online convex optimization. Foundations and Trends{\textregistered} in Machine Learning  \textbf{4}(2),  107--194 (2012)

\bibitem{shioura2017algorithms}
Shioura, A.: Algorithms for {L}-convex function minimization: {C}onnection between discrete convex analysis and other research fields. Journal of the Operations Research Society of Japan  \textbf{60}(3),  216--243 (2017)

\bibitem{pmlr-v206-tsuchiya23a}
Tsuchiya, T., Ito, S., Honda, J.: Further adaptive best-of-both-worlds algorithm for combinatorial semi-bandits. In: Proceedings of The 26th International Conference on Artificial Intelligence and Statistics. pp. 8117--8144. PMLR (2023)

\bibitem{zhang2021stochastic}
Zhang, H., Zheng, Z., Lavaei, J.: Stochastic $\mathrm{L}^{\natural}$-convex function minimization. Advances in Neural Information Processing Systems  \textbf{34},  13004--13018 (2021)

\end{thebibliography}
\end{document}